\documentclass[preprint,12pt]{elsarticle}

\usepackage{latexsym,amssymb,amsmath}
\usepackage{amsthm}
\usepackage{thmtools}
\usepackage{float}

\usepackage{tikz-cd}
\usepackage{cleveref}
\usepackage[T1]{fontenc}

\newtheorem{theo}{Theorem}

\newtheorem{prop}[theo]{Proposition}
\newtheorem{cor}[theo]{Corollary}

\crefname{theo}{theorem}{theorems}
\Crefname{theo}{Theorem}{Theorems}
\crefname{lem}{lemma}{lemmas}
\Crefname{lem}{Lemma}{Lemmas}
\crefname{prop}{proposition}{propositions}
\Crefname{prop}{Proposition}{Propositions}
\crefname{cor}{corollary}{corollaries}
\Crefname{cor}{Corollary}{Corollaries}

\crefname{figure}{figure}{figures}
\Crefname{figure}{Figure}{Figures}
\crefname{equation}{}{}
\Crefname{equation}{}{}

\journal{...}

\begin{document}

\begin{frontmatter}

\title{%
  Learning the parameters of a differential equation from its trajectory via the adjoint equation%
}

\author{%
    Imre Fekete\corref{cor1}%
}
\ead{imre.fekete@ttk.elte.hu}

\author{%
    Andr\'as Moln\'ar%
}

\author{%
    P\'eter L. Simon\\[0,3cm]
    \textit{Department of Applied Analysis and Computational Mathematics, Institute of Mathematics, Eötvös Loránd University, Hungary}\\[0,2cm]
\textit{MTA-ELTE Numerical Analysis and Large Networks Research
Group, Hungary}
}


\cortext[cor1]{Corresponding author}



\begin{abstract}
The paper contributes to strengthening the relation between machine learning and
the theory of differential equations.
In this context,
the inverse problem of fitting the parameters, and the initial condition of a
differential equation to some measurements constitutes a key issue.
The paper explores an abstraction that can be used to construct a family of loss functions with the aim of fitting the solution of an initial value problem to a set of discrete or continuous measurements.
It is shown, that an extension of the adjoint equation can be used to derive the gradient of the loss function as a continuous analogue of backpropagation in machine learning.
Numerical evidence is presented that under reasonably controlled circumstances the gradients obtained this way can be used in a gradient descent to fit the solution of an initial value problem to a set of continuous noisy measurements, and a set of discrete noisy measurements that are recorded at uncertain times.
\end{abstract}


\end{frontmatter}

\section{Introduction}
\label{section:introduction}
Machine learning has been connected to the field of differential equations recently, by observing that numerical time integrators resemble formulae used for residual neural networks \cite{RuthottoHaber,LuZhongLiDong}. This has led to the development of a significant number of new results appearing in several papers, some of which we now list as a non-exhaustive starting point for the interested reader \cite{Weinan,HaberRuthotto,DupontDoucetTeh,KimJiDengMaRackauckas}. In this paper, inspired by \cite{ChenRubanova}, we consider the problem of finding a differential equation, the solutions of which best fit a set of data.

The problem considered here can be formulated as follows. We are given
a set of time points $\mathcal{T} \subseteq [0,1]$, and a sample
from a trajectory of a differential equation evaluated at these points.
We remark that the choice of the unit interval is merely an aesthetic one, which
can be made without loss of generality.
This is typically either the time dependence of a trajectory component
$y:[0,1]\to \mathbb{R}$,
or a time series $y(\tau_1), y(\tau_2), \ldots , y(\tau_n)$ obtained from it.
The goal is to find an initial value problem, the solution of which fits the
given data.

More precisely, given a family of right hand sides parameterized by
a $k$-dimensional parameter $\theta \in \mathbb{R}^k$,
a $d$-dimensional initial condition $x_0 \subseteq \mathbb{R}^d$, and
a $1$-dimensional initial time $t_0 \in \mathbb{R}$,
we are looking for the best initial time, initial condition, parameter triple
$(t_0, x_0, \theta)$ in some search space
$\mathcal{S} \subseteq \mathbb{R} \times \mathbb{R}^d \times \mathbb{R}^k$.

That is, given the function
$f: \mathbb{R} \times \mathbb{R}^d \times \mathbb{R}^k \to \mathbb{R}^d$,
we consider the solution $x$ of the problem
\begin{equation}
\begin{cases}
  \dot x(t) &= f(t, x(t), \theta), \quad\quad\quad t_0 < t < t_0 + 1\\
  x(t_0) &= x_0, \label{equation:diffeq}
\end{cases}
\end{equation}

and try to find the value of $(t_0, x_0, \theta)$, for which the distance of the
functions $t \mapsto x(t)$ and $t \mapsto y(t - t_0)$ is minimal in some sense.

To this end, we employ a learning process, which first constructs a differentiable loss function
$\mathcal{L}\colon \mathcal{S} \to \mathbb{R}$,
then, given an initial guess for the triple $(t_0, x_0, \theta)$,
applies a gradient-descent based iterative method to minimize it.
Efficient calculation of the gradients used during the iteration is made
possible by the continuous backpropagation process based on the adjoint
equation \cite{ChenRubanova}.

As an illustrative example, the reader may have in mind the $d=1$ dimensional
case. Then two simple possible loss functions are the following. Given a discrete sample, we may let
\begin{subequations}\label{equation:1d-examples}
  \begin{equation}
\mathcal{L}(t_0, x_0, \theta) = \frac{1}{n} \sum_{j=1}^n (x_{(t_0, x_0, \theta)}(t_0 + \tau_j)-y(\tau_j))^2,\label{equation:1d-examples-discrete}
\end{equation}
while given the trajectory itself, we may pick
\begin{equation}
\mathcal{L}(t_0, x_0, \theta) = \int_{[0,1]} (x_{(t_0, x_0, \theta)}(t_0 +
\tau)-y(\tau))^2 \, d\tau,\label{equation:1d-examples-continuous}
\end{equation}
\end{subequations}
where we use the subscript $(t_0, x_0, \theta)$ to emphasize the solution's dependence on these parameters.

The paper is structured as follows.
In \Cref{section:general-approach}, we present the abstract approach,
construct the general loss function from building blocks,
and prove in \Cref{theorem:adjoint-equation}, that the adjoint equation yields the gradient of these.

Then, in \Cref{section:adjoint-equation-single-time-point}, and \Cref{section:adjoint-equation-multiple-time-points},
the adjoint equation is formulated, and the gradient of the general loss function is derived for the case of single,
and multiple time points, see \Cref{theorem:adjoint-equation-for-trajectory}.
In \Cref{section:application}, we turn to implementing the abstract approach.
In practice, to obtain the aforementioned gradient, one can solve the initial
value problem \Cref{equation:adjoint-ivp}, which presents the computable form of
the adjoint equation, and the suitable initial condition.
Lastly, in \Cref{section:numerical-examples} we show some numerical examples
illustrating the feasibility of the method.

The novelties in the paper are the abstract approach that enables us to treat the
discrete, and continuous cases together via a general loss function,
and a proof that an appropriately defined adjoint equation yields the
gradient of the general loss function.
This continuous form of backpropagation is presented here as a homotopy mapping
a function given at the output to a function acting at the input, see
Corollaries \ref{corollary:single} and \ref{corrolary:multiple}.
The numeric examples deal with continuous data that contains some spatial noise, and discrete data that contains some temporal and spatial noise.

\section{General approach}
\label{section:general-approach}
We will use the following standard notation for the solution that enables us to denote more clearly its dependence on the initial condition and on the parameters. Let $\phi(t, s, p, \theta)=x(t)$ denote the value of the solution of \Cref{equation:diffeq} at time $t$ satisfying the initial condition $x(s)=p$.
Then the initial value problem \Cref{equation:diffeq} takes the form
\begin{align*}
 \dot x_{(t_0, x_0, \theta)}(t) &= \partial_1 \phi(t, t_0, x_0, \theta) = f(t, \phi(t, t_0, x_0, \theta), \theta)
\end{align*}
for $t_0 < t < t_0 + 1$.
Moreover, we introduce the forward transfer operator family
$\varphi(\tau): \mathcal{S} \to \mathcal{S}$
by the formula
\begin{equation}
  \varphi(\tau)(s, p, \theta) = (\tau+s, \phi(\tau+s, s, p,\theta), \theta).
  \label{equation:varphidef}
\end{equation}
In words, $\varphi(\tau)$ advances the lifted dynamical system by time $\tau$.

The function $\varphi$ defines a dynamical system on the search space $\mathcal{S}$ and satisfies an autonomous differential equation, the right hand side of which is the lifted version of $f$, namely $F:\mathcal{S} \to \mathcal{S}$, defined as
\begin{align*}
  F(s, p, \theta) &= (1, f(s, p, \theta), 0),
\end{align*}
that is, the following proposition holds.
\begin{prop}
  \label{prop:dynsys}
The function $\varphi$ satisfies the group property $\varphi (t+\tau) = \varphi (t) \circ \varphi (\tau)$ and the autonomous differential equation
$$
\varphi '(\tau) = F \circ \varphi (\tau)
$$
for all $t$.
\end{prop}
\begin{proof}
The group property can be derived by using the group property of $\phi$ as follows.
\begin{align*}
\varphi (t) (\varphi (\tau)(s, p, \theta) ) &= \varphi (t)(\tau+s, \phi(\tau+s, s, p,\theta), \theta)\\
&=(t+\tau+s, \phi(t+\tau+s, \tau+s, \phi(\tau+s, s, p,\theta),\theta), \theta) \\
&= (t+\tau+s, \phi(t+\tau+s, s, p,\theta), \theta) =\varphi (t+\tau) (s, p, \theta).
\end{align*}
The differential equation can be obtained by differentiating \Cref{equation:varphidef} with respect to $\tau$.
\begin{equation*}
 \varphi'(\tau)(s, p, \theta)
 = (1, \partial_1\phi(s+\tau, s, p, \theta), 0)
 = (1, f(s+\tau, \phi(s+\tau, s, p, \theta), \theta), 0)
 = (F \circ \varphi(\tau))(s, p, \theta).
\end{equation*}
\end{proof}


We are now ready to construct the loss function.
The input of this function will be the triple $(t_0, x_0, \theta)$  including both the initial condition and the parameters. This triple determines the solution of the initial value problem \Cref{equation:diffeq} uniquely on $[t_0, t_0 + 1]$. The value of the loss function compares the measurement $y(\tau)$ to the state $\phi(t_0 + \tau, t_0, x_0, \theta)$ for some time instants $\tau\in [0,1]$.

To this end, we introduce the differentiable function
$h(\tau): \mathcal{S} \to \mathbb{R}$,
that maps the state triple at time $t_0 + \tau$ to a scalar representing the error at this time.

One of the most typical error functions is the square of the difference,
that is used in the $d=1$ dimensional cases
\Cref{equation:1d-examples-discrete,equation:1d-examples-continuous} of
\Cref{section:introduction}. In that case, the function $h(\tau)$ takes the form of
\[
  h(\tau)(s, p, \theta) = (p - y(\tau))^2.
\]

To turn this into a function of the initial state, we compose it from the right by the function $\varphi(\tau)$, which advances the state by time $\tau$.
The result is the function
\[
  h(\tau) \circ \varphi(\tau): \mathcal{S} \to \mathbb{R}.
\]
In the case of the simple squared difference of \Cref{equation:1d-examples-discrete,equation:1d-examples-continuous}, we get 
\[
\left( h(\tau) \circ \varphi(\tau) \right) ( t_0, x_0, \theta) = (\phi(t_0 + \tau, t_0, x_0, \theta) - y(\tau))^2.
\]

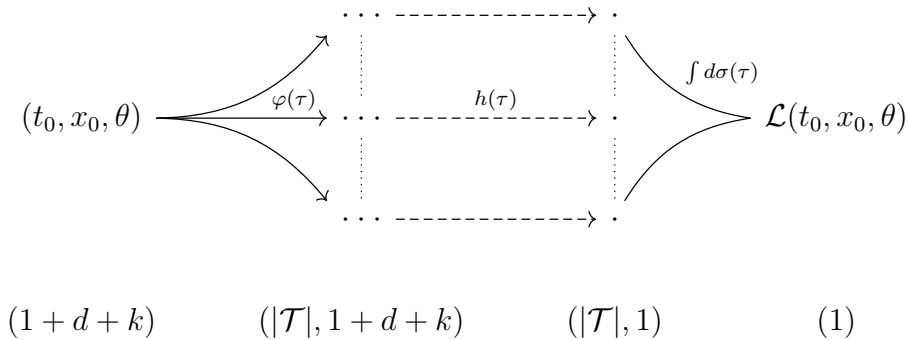
\begin{figure}[H]
  \begin{center}
    \begin{tikzcd}
      {}&{\cdot\cdot\cdot\arrow[d, no head, dotted]} \arrow[r, dashed] &
      \cdot \arrow[d, no head, dotted]
      \arrow[rd, bend right, no head, "\int d\sigma(\tau)", end anchor=west,
      bend right=25]&
   &
   \\
   (t_0, x_0, \theta)
   \arrow[start anchor=east, ru, bend right=25, end anchor=south west]
   \arrow[start anchor=east, rd, bend left=25,  end anchor=north west]
   \arrow[start anchor=east, r, "\phantom{------}\varphi(\tau)"]
   &{\cdot\cdot\cdot\arrow[d, no head, dotted]} \arrow[r, dashed, "h(\tau)"] &
   \cdot \arrow[d, no head, dotted]                                   &
   \mathcal{L}(t_0, x_0, \theta) &
   \\
   {}&{\cdot\cdot\cdot} \arrow[r, dashed]                                               &
   {\cdot} \arrow[ru, bend left, no head, end anchor=west, bend left=25] & &
   \\
   (1 + d + k) & (|\mathcal{T}|, 1+d+k) & (|\mathcal{T}|, 1) & (1)
 \end{tikzcd}
\end{center}
\caption{The loss function $\mathcal{L}$, which, in words,
  for each time $0 \leq \tau \leq 1$,
  transfers the initial state triple $(t_0, x_0, \theta)$ forward by time $\tau$,
  assigns a scalar score to the resulting state triple using $h(\tau)$,
  and lastly aggregates these scores by integrating over [0, 1] with respect to
  the measure $\sigma$. The bottom row lists the dimensions, and shapes of the
  objects encountered, in a form related to implementation. These are, from left to right: a (row) vector, a matrix
  with the same number of columns, and $|\mathcal{T}|$ rows, that is, one for each time instant; a column vector with the same number of rows, and lastly a scalar. }
\label{figure:loss-function-explained}
\end{figure}

If we  want to compare the solution to the measurement at several time instants $\tau \in [0, 1]$, and then aggregate the resulting differences, then we take a probability measure $\sigma$ on $[0,1]$ that is concentrated to those time instants and integrate the point-wise error $h(\tau) \circ \varphi(\tau)$ with respect to this measure, leading to the general definition of the loss function as follows
\begin{equation}
\mathcal{L} = \int_{[0, 1]} h(\tau) \circ \varphi(\tau) \, \, d\sigma(\tau). \label{equation:loss-function}
\end{equation}
To emphasize the arguments of the loss function, this definition can be written in the form
\[
\mathcal{L}( t_0, x_0, \theta) = \int_{[0, 1]} (h(\tau) \circ \varphi(\tau))( t_0, x_0, \theta) \, \, d\sigma(\tau).
\]
We visualize the general loss function in \Cref{figure:loss-function-explained}.


The goal of the learning process is to find a minimum of the loss function in
the search space, i.e. to find the optimal values of the initial condition $(t_0,x_0)$ and the parameter $\theta$.
To this end, the efficient calculation of the gradient of the loss function, denoted by $\mathcal{L}'$, is needed.
Equation \Cref{equation:loss-function} shows that this gradient can be obtained from the derivative $(h(\tau) \circ \varphi(\tau))'$.
It turns out that computing this derivative is numerically demanding, hence an alternative route using the so-called adjoint equations has been developed, see e.g. \cite{ChenRubanova}.
Below we show a general derivation of this equation and a new proof for the fact that the gradient of the loss function can be obtained from the adjoint equation.

The main idea of this general approach is that calculating $h(\tau)' \circ \varphi(\tau)$ is relatively easy, and it is connected to the desired derivative $(h(\tau) \circ \varphi(\tau))'$ by a differential equation, the adjoint equation.

In other words, we show that there exists a differential equation, such that its solution acts as a continuous transformation between
the functions $(h(\tau) \circ \varphi(\tau))'$ and $h(\tau)' \circ \varphi(\tau)$, much like a homotopy mapping one curve to another.

Indeed, given a time $0 \leq t \leq \tau$, let us define
\[
  \Lambda(\tau, t) = h(\tau) \circ \varphi(\tau - t),
\]
and use the group property of $\varphi$ to split the map $h (\tau) \circ \varphi(\tau)$ as
\[
 h (\tau) \circ \varphi(\tau) = h(\tau) \circ \varphi(\tau - t) \circ \varphi(t) =  \Lambda(\tau, t) \circ \varphi(t).
\]
Now, we introduce the desired homotopy $\lambda(\tau, t)$ as follows
\[
  \lambda(\tau, t) =
  (h(\tau) \circ \varphi(\tau - t))' \circ \varphi(t) =
  \Lambda(\tau, t)' \circ \varphi(t).
\]
Clearly, then $\lambda(\tau, \tau) = h(\tau)' \circ \varphi(\tau)$, and
$\lambda(\tau, 0) = (h(\tau) \circ \varphi(\tau))'$ hold, i.e. $\lambda$
connects the two mappings.
The time evolution of $\lambda$, that is the function $t \mapsto \lambda(\tau, t)$ satisfies a differential equation, that is generally called the adjoint equation.
This is the statement of the following theorem.

\begin{theo}
  \label{theorem:adjoint-equation}
  The function $\lambda(\tau, \cdot)$ satisfies the differential equation
\begin{equation}
\partial_t \lambda(\tau, t) = - \lambda(\tau, t) \cdot (F' \circ \varphi(t))
   \quad \quad \mbox{ for } \quad  0 < t < \tau \leq 1 .\label{equation:adjointeq}
\end{equation}
\end{theo}
\begin{proof}
  By the group property, and the chain rule, we have that
    \begin{align*}
      \Lambda(\tau, t) &= \Lambda(\tau, t+s) \circ \varphi(s), \\
      \Lambda(\tau, t)' &= (\Lambda(\tau, t+s)' \circ \varphi(s)) \cdot \varphi(s)'.
    \end{align*}
  Applying this to $\lambda$, we get that
  \begin{align*}
  \lambda(\tau, t) &= \Lambda(\tau, t)' \circ \varphi(t) \\
                   &= (\Lambda(\tau, t + s)' \circ \varphi(s) \circ \varphi(t)) \cdot (\varphi(s)' \circ \varphi(t)) \\
  &= \lambda(\tau, t+s) \cdot (\varphi(s)' \circ \varphi(t)).
\intertext{
  Now we take the derivative with respect to $s$, and substitute $s=0$.
}
    0 &= \left. \partial_t \lambda(\tau, t+s) \cdot (\varphi(s)' \circ \varphi(t))
        + \lambda(\tau, t+s) \cdot \frac{d}{ds}\left(\varphi(s)' \circ \varphi(t)\right) \right|_{s=0}
    \\&= \partial_t \lambda(\tau, t) + \lambda(\tau, t) \cdot (\varphi'(0)' \circ \varphi(t))
    \\&= \partial_t \lambda(\tau, t) + \lambda(\tau, t) \cdot (F' \circ \varphi(t)),
\intertext{
  where the last line uses
}
    \left.\frac{d}{d\tau}(\varphi(\tau)')\right|_{\tau=0} &= (\varphi'(0))' = (F \circ \varphi(0))' = F'.
    \end{align*}
\end{proof}

To summarize, the general approach is to solve the the differential equation
\Cref{equation:diffeq}, then the gradient of the loss function is obtained by
solving the adjoint equation backward, from $t=t_0+\tau$ to $t=t_0$.
So far we have obtained the derivative $(h(\tau) \circ \varphi(\tau))'$.
In the next two sections, we present how to get the gradient of the loss function when we have only a single time point, i.e. the probability measure is concentrated to a single point, and when we have several time instants.

\section{The case of a single time point}
\label{section:adjoint-equation-single-time-point}
Let us first consider the case of a single measurement at a fixed time $\tau$.
This corresponds to the case where $\sigma$ is concentrated on the single time instant $\tau$.
Then, the loss function is simply
$
  h(\tau) \circ \varphi(\tau)
  $, which acts on $\mathcal{S}$ by the formula
\begin{equation}
  \mathcal{L}(t_0, x_0, \theta) = h(\tau)(t_0 + \tau, \phi(t_0+\tau, t_0, x_0, \theta), \theta).\label{equation:loss-func-single}
\end{equation}
For the sake of brevity, and exploiting that $\tau$ is fixed now, we introduce the functions $\bar h=h(\tau)$, and $\bar \varphi = \varphi(\tau)$, and we let $\xi_0 = (t_0, x_0, \theta)$. Using these notations the loss function can be written as
\[
  \mathcal{L}(\xi_0) = \bar h( \bar \varphi(\xi_0)).
\]

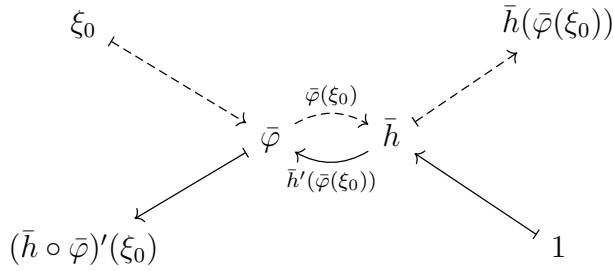
\begin{figure}[H]
  \begin{center}
    \begin{tikzcd}
     \xi_0 \arrow[dr, maps to, dashed]&&&{\bar h(\bar \varphi(\xi_0))}
     \\
     {} & \bar \varphi
     \arrow[r, bend left, "\bar \varphi(\xi_0)", dashed]
     \arrow[ld, maps to]
     &{\bar h} \arrow[ru, maps to, dashed] \arrow[l, bend left, "\bar h'(\bar \varphi(\xi_0))"]
     \\
     {(\bar h\circ \bar \varphi)'(\xi_0)}&&&{1}\arrow[lu, maps to]
   \end{tikzcd}
  \end{center}
  \caption{The forward, and the backward pass in the case of a single time point $\tau$.
    The arrows representing the former are dashed.
    During the forward pass we start from $\xi_0$ and calculate $\bar \varphi(\xi_0)$, then $\bar h( \bar \varphi(\xi_0))$.
    During the backward pass we take these values, and starting from $1 = \operatorname{id}'(\bar h( \bar \varphi(\xi_0)))$,
    we calculate $\bar h'(\bar \varphi(\xi_0))$, and lastly $(\bar h\circ \bar \varphi)'(\xi_0)$.
  }
  \label{figure:loss-function-single-point-case}
\end{figure}

We are interested in calculating the gradient of this function using backpropagation,
summarized in \Cref{figure:loss-function-single-point-case}.

We note, again, that in the simple case when $\bar h (s,p,\theta)=(p-y(\tau))^2$, the loss function takes the form
\[
\mathcal{L}(t_0, x_0, \theta) = (\phi(t_0 + \tau, t_0, x_0, \theta) - y(\tau))^2 .
\]

Based on the result of the previous section, the gradient of the loss function can be calculated as follows.
\begin{cor}
  Let the loss function be given by \Cref{equation:loss-func-single}.
  Then its gradient can be obtained as $ \mathcal{L} ' =\lambda(\tau, 0)$, where $\lambda(\tau, \cdot)$ is the solution of the adjoint equation \Cref{equation:adjointeq}, solving it backward starting from the initial condition $\lambda(\tau, \tau) = \bar h' \circ \bar \varphi$ with $\bar h=h(\tau)$, and $\bar \varphi = \varphi(\tau)$.
\label{corollary:single}
\end{cor}

\section{The case of multiple time points}
\label{section:adjoint-equation-multiple-time-points}

Similarly to the single point case, we would like to find a way to transform the various $\lambda(\tau,\tau) = h(\tau)' \circ \varphi(\tau)$ functions, possibly scaled values of which are obtained during backpropagation, into the derivative of the loss function
\Cref{equation:loss-function}, that is, into $\mathcal{L}'$.

Given a $0 \leq t \leq 1$, let us consider how the loss function depends on the state at time $t$.
During the forward pass, that is, the evaluation of the loss function $\mathcal{L}$, the initial value problem \Cref{equation:diffeq} is solved forward in time.
This implies that the aforementioned state affects the states at later times, that is, those at time $\tau$ for all $t \leq \tau \leq 1$.

The effect is the following.
First, the state is carried to time $\tau$ via $\varphi(\tau - t)$, then the resulting state is fed into $h(\tau)$, yielding the partial loss value belonging to time $\tau$. Therefore, we form the composition of these two functions,
\[
h(\tau) \circ \varphi(\tau - t)
\]
for each $t \leq \tau \leq 1$,
and aggregate the results using the measure $\sigma$ to get the function
\begin{align*}
  L(t) &= \int_{[0,1]} \mathbb{I}(t \leq \tau) \cdot h(\tau) \circ \varphi(\tau - t) \,d \sigma(\tau),
\end{align*}
which can be seen to be the $\tau-$aggregated version of $\Lambda(\tau, t)$.
This becomes a proper loss function, in the sense that it will take the initial state to some loss value, if we compose it from the right by $\varphi(t)$.
Indeed,
\[
  L(t) \circ \varphi(t)
\]
is a family of loss functions that measure the loss encountered on the interval
$[t, 1]$.
Using that $\varphi(0) $ is the identity, equation \Cref{equation:loss-function} yields $L(0)= \mathcal{L}$.
%
%

We may now proceed analogously to the single point case, and define
\begin{align*}
  l(t) &= L(t)' \circ \varphi(t)
         \\
  &= \int_{[0,1]} \mathbb{I}(t \leq \tau) \cdot (h(\tau) \circ \varphi(\tau
  - t))' \circ \varphi(t)\,d \sigma(\tau)
         \\
        &= \int_{[0,1]} \mathbb{I}(t \leq \tau) \cdot \lambda(\tau, t) \,d \sigma(\tau),
\intertext{
the $\tau-$aggregated version of $\lambda(\tau, t)$,
which will act as the
transformation between the functions
}
  l(0) &= \mathcal{L}', \\
  l(1) &= \sigma(\{1\}) \cdot h(1)'\circ \varphi(1).
\end{align*}
Let us describe now the time evolution of $l$.
The case of the continuous and the discrete sample can be treated together by assuming that
$\sigma$ decomposes into the sum of an absolutely continuous and a discrete part,
that is $\sigma = \sigma_c + \sigma_d$ with Radon-Nikodym derivatives $\rho_c$ and $\rho_d$.
Then we have that
\begin{equation}
  l(t) =  \int_{0}^1 \mathbb{I}(t \leq \tau) \cdot \lambda(\tau, t) \cdot \rho_c(\tau)\,d \tau
  + \sum_{j=1}^n \mathbb{I}(t \leq \tau_j) \cdot \lambda(\tau_j, t)\cdot \rho_d(\tau_j),
  \label{equation:l(t)}
\end{equation}
and the time evolution of this family is given by the following theorem.
\begin{theo}
  \label{theorem:adjoint-equation-for-trajectory}
  \begin{equation}
    l'(t) = - \lambda(t, t) \cdot \rho_c(t) - \sum_{j = 1}^n \lambda(\tau_j, \tau_j)\cdot \rho_d(\tau_j) \cdot \delta_{\{\tau_j\}} - l(t) \cdot (F'\circ \varphi(t))
    \quad\quad\quad 0 < t < 1 \label{equation:adjointeq-trajectory}
  \end{equation}
\end{theo}
\begin{proof}
  The idea of the proof is to differentiate \Cref{equation:l(t)}, and apply \Cref{theorem:adjoint-equation}.
  For the continuous part, we use the Leibniz rule.
  \begin{align*}
    l'(t) &= -\lambda(t, t) \rho_c(t) + \int_{t}^1 \partial_t\lambda(\tau, t) \rho_c(\tau) \,d \tau
    \\
          &- \sum_{j = 1}^n \lambda(\tau_j, \tau_j)\rho_d(\tau_j) \delta_{\{\tau_j\}} +  \sum_{j=1}^n \mathbb{I}(t \leq \tau_j) \cdot \partial_t \lambda(\tau_j, t)\rho_d(\tau_j)
    \\
          &= -\lambda(t, t) \rho_c(t) - \int_{t}^1 \lambda(\tau, t)\cdot (F'\circ \varphi(t)) \rho_c(\tau) \,d \tau
    \\
          &- \sum_{j = 1}^n \lambda(\tau_j, \tau_j)\rho_d(\tau_j) \delta_{\{\tau_j\}} -  \sum_{j=1}^n \mathbb{I}(t \leq \tau_j) \cdot \lambda(\tau, t)\cdot (F'\circ \varphi(t))\rho_d(\tau_j)
    \\
          &= -\lambda(t, t) \rho_c(t) - \sum_{j = 1}^n \lambda(\tau_j, \tau_j)\rho_d(\tau_j) \delta_{\{\tau_j\}}
    \\
          & - \left( \int_{t}^1 \lambda(\tau, t)\rho_c(\tau) \,d \tau
            +  \sum_{j=1}^n \mathbb{I}(t \leq \tau_j) \cdot \lambda(\tau, t) \rho_d(\tau_j) \right) \cdot (F'\circ \varphi(t))
    \\
          &= - \lambda(t, t) \cdot \rho_c(t) - \sum_{j = 1}^n \lambda(\tau_j, \tau_j)\cdot \rho_d(\tau_j) \cdot \delta_{\{\tau_j\}} - l(t) \cdot (F'\circ \varphi(t))
  \end{align*}
\end{proof}
We take a moment to underline yet again that $\lambda(t, t) =
h(t)' \circ  \varphi(t)$, and that $\lambda(t, t)$ are functions from
which we obtain values during backpropagation.

\begin{cor}
  Consider the general loss function \Cref{equation:loss-function}. Its
  gradient is $ \mathcal{L} ' = l(0)$, where $l$ is the solution of the
  adjoint equation \Cref{equation:adjointeq-trajectory}, which we solve backward
  in time starting from the initial condition $l(1) = \sigma(\{1\}) \cdot
  h(1)'\circ \varphi(1)$. \label{corrolary:multiple}
\end{cor}

\section{Application of the general theory}
\label{section:application}
In this section, we turn to the application of the general theory presented above. As the initial setting, we are given the input to $\mathcal{L}$, namely the triple $(t_0, x_0, \theta)$.

During the forward pass, the initial value problem \Cref{equation:diffeq} is solved to produce a solution $x_{(t_0, x_0, \theta)}$, which we denote simply by $x$, for the sake of brevity. This is then fed into
the functions $h(\tau)$ point-wise, the results of which are aggregated via integration by the measure $\sigma$ on $[0,1]$.

During the backward pass, we use $x$, a result of the forward pass, and solve another initial value problem backwards in time to backpropagate the gradient obtained in the form of a function $g$.
We note that if we have a finite number of time points, then $g$ is really just a finite dimensional vector.

\subsection{The case of a single time point}

First, we illustrate how to apply the general theory in the case of a single
time point $\tau$. To simplify matters as much as possible, we consider a
differential equation with a $d=1$ dimensional phase space and a $k=1$
dimensional parameter. Moreover, we pick the squared difference error function
$h(\tau)(s,p,\theta)=(p-y(\tau))^2$. In this case, the loss function maps
$\mathbb{R}^3$ to $\mathbb{R}$ following the formula
\[
\mathcal{L}(t_0, x_0, \theta) = (\phi(t_0 + \tau, t_0, x_0, \theta) - y(\tau))^2 ,
\]
which is consistent with \Cref{equation:1d-examples-discrete}, assuming $n=1$ observation(s).

According to \Cref{corollary:single}, the derivative of the loss function is $ \mathcal{L} ' =\lambda(\tau, 0)$, where $\lambda(\tau, \cdot)$ is the solution of the adjoint equation \Cref{equation:adjointeq} satisfying the initial condition $\lambda(\tau, \tau) = h(\tau)' \circ\varphi(\tau)$.

The adjoint equation \Cref{equation:adjointeq} is in a functional form. Applying
both the left and the right-hand-sides to a point $(t_0, x_0, \theta)$ leads to
a linear system of three differential equations. Let us now expand on these.
First, we introduce the function that is going to satisfy this linear
differential equation as
\[
  (a_1(t),a_2(t),a_3(t))=a(t)= \lambda(\tau, t) (t_0, x_0, \theta),
\]
where components $a_i$ are now real-valued functions.

Then the adjoint equation itself is the non-autonomous linear differential equation of the form
\[\dot a(t)= -a(t) A(t),\] where the coefficient
matrix is $A(t)= F'(\varphi(t)(t_0, x_0, \theta))$. Elaborating on this, we note
that since $\varphi(t)(t_0, x_0, \theta)=( t_0+t, x(t_0 + t) , \theta)$, where $x(t_0 + t) = \phi(t_0+t,t_0,x_0,\theta)$, and
\[
F'(s,p,\theta) =\left(
                  \begin{array}{ccc}
                    0 & 0 & 0 \\
                    \partial_1 f(s, p, \theta) & \partial_2 f(s, p,\theta) &
                                                                             \partial_{3} f(s, p,\theta) \\
                    0 & 0 & 0 \\
                  \end{array}
                \right),
\]
we have that
\[
A(t) =\left(
                  \begin{array}{ccc}
                    0 & 0 & 0
   \\
                    \partial_1 f(t_0 + t, x(t_0+t), \theta) &
                    \partial_2 f(t_0 + t, x(t_0 + t),\theta) &
                    \partial_{3} f(t_0 + t, x(t_0 + t),\theta)
   \\
                    0 & 0 & 0
                  \end{array}
                \right).
\]
Therefore, multiplication leads us to the expanded version of the adjoint equation,
\begin{align}
  \dot a_1(t) &= -a_2(t) \partial_1 f(t_0 + t, x(t_0 + t), \theta), \label{eq:deqa1}
   \\
  \dot a_2(t) &= -a_2(t) \partial_2 f(t_0 + t, x(t_0 + t),\theta), \label{eq:deqa2}
  \\
  \dot a_3(t) &= -a_2(t) \partial_{3} f(t_0 + t, x(t_0 + t),\theta) . \label{eq:deqa3}
\end{align}
Thus, we need to solve the second equation for $a_2$, first, and then $a_1$ and
$a_3$ can be obtained by simple integration.

Lastly, we derive the initial conditions for the unknown functions $a_i$. The
abstract initial condition takes the form $\lambda(\tau, \tau) = h(\tau)'
\circ\varphi(\tau)$, and we have that $a(\tau)= \lambda(\tau, \tau) (t_0, x_0, \theta)$. Differentiating $h(\tau)(s,p,\theta)=(p-y(\tau))^2$ yields
\[
h(\tau)'(s,p,\theta)=( 0, 2(p-y(\tau)), 0) .
\]
Using $\varphi(\tau)(t_0, x_0, \theta)=( t_0+\tau, x(t_0 + \tau) , \theta)$, we obtain
\[
a(\tau)= h(\tau)' (\varphi(\tau)(t_0, x_0, \theta) ) =  ( 0, 2(x(t_0 + \tau)-y(\tau)) , 0),
\]
leading to the initial condition
\begin{equation}
a_1(\tau)= 0, \qquad a_2(\tau)= 2(x(t_0 + \tau)-y(\tau)),  \qquad a_3(\tau)= 0 . \label{eq:icai}
\end{equation}
Thus, the gradient of the loss function can be obtained as
\[
\mathcal{L}'(t_0, x_0, \theta) = a(0),
\]
where $a(t)=(a_1(t),a_2(t),a_3(t))$ is the solution of system \eqref{eq:deqa1}-\eqref{eq:deqa3} subject to the initial condition \eqref{eq:icai}.

For the interested Reader, it might be useful to consider the case $f(p,\theta)=p\theta$, when system \eqref{eq:deqa1}-\eqref{eq:deqa3} can be solved analytically as
\[
a(t)= 2(\mbox{e}^{\theta \tau} x_0 - y(\tau)) (0, \mbox{e}^{\theta (\tau -t)} , \mbox{e}^{\theta \tau} x_0 (\tau -t)),
\]
leading to
\[
\mathcal{L}'(t_0, x_0, \theta) = a(0) = 2(\mbox{e}^{\theta \tau} x_0 - y(\tau)) (0, \mbox{e}^{\theta \tau} , \tau  \mbox{e}^{\theta \tau} x_0 ) .
\]
In this special case, the gradient of the loss function can also simply be obtained by direct differentiation of
\[
\mathcal{L}(t_0, x_0, \theta) = (\mbox{e}^{\theta \tau}x_0 - y(\tau))^2 .
\]

\subsection{The case of multiple time points}

The case of multiple time points can be treated similarly to the single point
case, seen in the previous
subsection.

We start by considering the general loss function $\mathcal{L}$ as
defined in \Cref{equation:loss-function}.
According to \Cref{corollary:single}, its derivative is calculable as $\mathcal{L}' = l(0)$, where $l$ is the solution of the adjoint equation
\Cref{equation:adjointeq-trajectory}, satisfying the initial condition $l(1) = \sigma(\{1\}) \cdot h(1)'\circ \varphi(1)$.

We now take \Cref{equation:adjointeq-trajectory} in its functional
form, and apply its functions to the input triple $(t_0, x_0, \theta)$.
Given a $t$ from the unit interval, the three functions that we need to evaluate
are $l(t), \lambda(t, t)$, and $F'\circ \varphi(t)$.
In doing so, we will freely use that $\varphi(\tau)(t_0, x_0, \theta)=( t_0+\tau, x(t_0 + \tau) , \theta)$.
We start with $l(t)$, and define the function that is to satisfy the adjoint equation as
\begin{align*}
  a(t) &= (a_1(t), a_2(t), a_3(t)) = l(t)(t_0, x_0, \theta) &\in \mathbb{R}^{1+d+k}.
  \intertext{
Then, we consider source term $\lambda(t, t) = h(t)' \circ \varphi(t)$, which might be considered the input gradient during the backpropagation step, and define the corresponding function
  }
  g(t) &= \left( h(t)' \circ \varphi(t)\right)(t_0, x_0, \theta) = h(t)'(t_0 + t, x(t_0 + t), \theta) &\in \mathbb{R}^{1+d+k}.
\intertext{
 Then, we mimic the previous subsection and let
}
  A(t) &=\left(F' \circ \varphi(t)\right)(t_0, x_0, \theta) = F'(t_0 + t, x(t_0 + t), \theta)
  &\in \mathbb{R}^{(1 + d + k)\times(1+d+k)}.
  \intertext{
    Lastly, we define
  }
  J(t) &= f'(t_0 + t, x(t_0 + t), \theta) &\in \mathbb{R}^{d \times (1+d+k)},
\end{align*}
and note that
\[
a \cdot A =
\begin{bmatrix}
  a_1& a_2& a_3
\end{bmatrix}
\cdot
\begin{bmatrix}
  0 \\ J \\ 0
\end{bmatrix} =
a_2 \cdot J.
\]

Still following \Cref{theorem:adjoint-equation-for-trajectory}, we are ready to state the initial value problem to be solved backward in time.
Indeed, we plug in the recently defined functions to get
\begin{equation}
  \begin{cases}
    \dot a(t) &= - g(t)\rho_c(t) - \sum\limits_{j = 1}^n g(\tau_j) \rho_d(\tau_j)\delta_{\{\tau_j\}}
    - a_2(t) \cdot J(t), \quad\quad\quad 0 < t < 1 \\
    a(1) &= \phantom{-}g(1)\rho_d(1), \label{equation:adjoint-ivp}
  \end{cases}
\end{equation}
where the initial value follows from the formula
\[
  a(1) = \sigma(\{1\})\cdot\left(h(1)'\circ\varphi(1)\right)(t_0, x_0, \theta) =
  \rho_d(1) \cdot g(1),
\]
where we have used that $\sigma = \sigma_d + \sigma_c$, and $\sigma_c(\{1\}) =
0$ by its absolute continuity.

To summarize, given the values $\{g(t): t \in \mathcal{T}\}$, the gradient of the loss
function can be obtained as
\[
\mathcal{L}'(t_0, x_0, \theta) = a(0),
\]
where $a$ is the solution of the initial value problem \Cref{equation:adjoint-ivp}.

We take a moment to state that the $i$th component of \Cref{equation:adjoint-ivp} for $i=1,2,3$ is
\begin{equation*}
  \begin{cases}
  \dot a_i(t) &= -g_i(t) \rho_c(t) - \sum\limits_{j=1}^n g_i(\tau_j) \rho_d(\tau_j)\delta_{\{\tau_j\}} -a_2(t) \partial_i f(t_0 + t, x(t_0 + t), \theta), \qquad 0 < t < 1
  \\
  a_i(1) &= \phantom{-}g_i(1)\rho_d(1),
  \end{cases}
\end{equation*}
and we note that it involves a nontrivial differential equation only for $i=2$,
therefore having solved that first, the rest of the components $a_1,$ and $a_3$
may be found by integration.

We note that
using a discrete set of observations in a
continuous world has its price, namely the Dirac delta
terms $\delta_{\{ \tau_j \}}$ mean that that $a$ has jumps of possibly nonzero magnitude at times $\tau_j$.
In practice, this means that the numerical algorithm used to solve problem
\Cref{equation:adjoint-ivp} has to be able to introduce artificial bumps in the
solution it is producing.
Alternatively, we may introduce the bumps by solving initial value problems on each sub interval $[1, \tau_{n}], \ldots, [\tau_{j}, \tau_{j-1}], \ldots
[\tau_{1}, 0]$, and bumping the solution $a$ through the initial conditions.

To make the latter argument more precise, we
firstly let $\tau_{n+1} = 1$, and $ \tau_0 = 0$, without
introducing new time instants, and define $a^{n+1} \equiv 0$.
Then, for each $j=n, \ldots, 0$, we recursively introduce a sequence of functions
\[
  a^j:[\tau_{j+1}, \tau_j] \to \mathbb{R}^{1+d+k},\qquad
\]
as the solutions to the sequence of initial value problems

\begin{equation}
  \begin{cases}
    \dot a^j(t) &= - g(t)\rho_c(t)
    - a^j_2(t) \cdot J(t), \quad\quad\quad \tau_j < t < \tau_{j+1} \\
    a^j(\tau_{j+1}) &= \phantom{-}g(\tau_{j+1}) \rho_d(\tau_{j+1}) + a^{j+1}(\tau_{j+1}),
    \label{equation:adjoint-ivp-sequence}
\end{cases}
\end{equation}
solving all of which in succession, we arrive at $a^0(\tau_0) = a^0(0) =
\mathcal{L}'(t_0, x_0, \theta)$.
We note that the $g(\tau_{j+1})\rho_d(\tau_{j+1})$ terms get added with a
positive sign, since a jump in forward time becomes the same jump, but negated, when looking at it in reversed time.

Lastly, we underline two important special cases. The first assumes that
continuous data is available on the whole unit interval, that is, when $y(\tau)$ is defined for
each $\tau$ from $[0,1]$.
We do not wish to highlight any single time instant in particular, therefore
we let $\rho_d \equiv 0$, and we set the continuous weights to be uniform, that
is, $\rho_c \equiv 1$. In other words, $\sigma$ is the Lebesgue-measure on $[0, 1]$. In this
case, the loss function is
\begin{equation*}
  \mathcal{L}(t_0, x_0, \theta) = \int_0^1 h(\tau)(t_0 + \tau, x(t_0 +
  \tau), \theta) \,\, d\tau,
\end{equation*}
and \Cref{equation:adjoint-ivp} becomes
\begin{equation}
  \begin{cases}
    \dot a(t) &= - g(t) - a_2(t) \cdot J(t), \quad\quad\quad 0 < t < 1 \\
    a(1) &= \phantom{-}0,
  \end{cases}
  \label{equation:adjoint-ivp-continuous}
\end{equation}
since $\rho_d \equiv 0$.

The second assumes that we have a single observation at time $\tau$.
In this case, $\sigma$ is concentrated on $\tau$, that is, the continuous part is zero,
$\rho_c \equiv 0$, while the discrete part is zero everywhere except at $\tau$,
where $\rho_d(\tau) = 1$.
We can consider three cases based on the value of $\tau \in [0, 1]$.
If $\tau = 0$, then there is no need to solve any initial value problem. If
$\tau = 1$, then
\Cref{equation:adjoint-ivp} becomes
\begin{equation*}
  \begin{cases}
    \dot a(t) &= - a_2(t) \cdot J(t), \quad\quad\quad 0 < t < 1 \\
    a(1) &= \phantom{-}g(1), 
  \end{cases}
\end{equation*}
where the right hand side doesn't show the Dirac delta term that sits at $\tau = 1$,
since it is outside of the interval where this differential equation is solved.
This is a terse version of the single point case outlined in the previous subsection.
If $0 < \tau < 1$, then
\Cref{equation:adjoint-ivp} becomes
\begin{equation*}
  \begin{cases}
    \dot a(t) &=  - g(\tau)\delta_{\{\tau\}} - a_2(t) \cdot J(t), \quad\quad\quad 0 < t < 1 \\
    a(1) &= \phantom{-}0, 
  \end{cases}
\end{equation*}
which is a homogeneous linear system on $(\tau, 1)$, and consequently, its
solution there is zero, because of the initial condition $a(1) = 0$.
At time $\tau$, $a$ has a jump of $g(\tau)$, and from that point, the
homogeneous differential equation can transfer the now non-zero state to
something other than zero. This process amounts to the solution of the initial value problem
\begin{equation*}
  \begin{cases}
    \dot a(t) &=  - a_2(t) \cdot J(t), \quad\quad\quad \tau > t > 0 \\
    a(\tau) &= \phantom{-}g(\tau), 
  \end{cases}
\end{equation*}
which is, again, what the treatment of the single point case of the previous
subsection predicted.


\section{Numerical experiments}
\label{section:numerical-examples}
In this section, we present the results of numerical experiments
as evidence in support of
\Cref{theorem:adjoint-equation-for-trajectory}.
We demonstrate that a gradient descent that obtains the necessary gradients via
\Cref{equation:adjoint-ivp-continuous} as outlined in this paper is able to lessen small perturbations in an optimal parameter triple $\xi_0 = (t_0, x_0, \theta)$.

The experiments proceed as follows.
To obtain our input data we solve an initial value problem
\Cref{equation:diffeq} parameterized by
$\xi_0$, and sample the first component of the resulting trajectory.
We consider two cases.

In the first, continuous case, we assume that the
entirety of this component is available to the optimization process.
To mimic measurement errors, each time this component is evaluated, the result
contains an additive error term that is normally distributed. In this case, the function
family $h$ is the square of the difference between the first component of the state of the
dynamical system and the sample $y$.

In the second, discrete case, we uniformly divide the unit interval into
subintervals. We then generate a discrete sample by considering the input data
of the previous case and sampling it at a time instant from each subinterval,
where these time instants are drawn from truncated normal distributions that are
centered at the intervals' midpoints. Our $y$ input data will then be a piecewise constant
function, which takes the sampled value on each subinterval.
We modify the $h$ of the continuous case by multiplying it with a weight
function, which is, on each subinterval, the probability density function of the
time instant where the trajectory component has been sampled.

Then we construct the computational graph, or loss function, using our input
data $y$, the vector field of the initial value problem $f$, and the loss
function components $h$.
Lastly, we apply a small random normal perturbation to the true parameter triple
$\xi_0$, and initiate a gradient descent starting from the perturbed triple, in
order to reduce the loss value.

As initial value problems, we consider
the SI model with a fixed population of $10$
\begin{equation}
  \begin{cases}
    \dot S &= - \frac{\beta IS}{10} \\
    \dot I &= \phantom{-} \frac{\beta IS}{10} - \gamma I
    \label{equation:sir-ivp}
  \end{cases}
  \quad\quad\quad\quad
  t_0 = 0
  \quad\quad
  \begin{cases}
    S(t_0) &= 9 \\
    I(t_0) &= \frac{1}{2}
  \end{cases}
  \quad\quad
  \begin{bmatrix}
    \beta & \gamma
  \end{bmatrix}
  =
  \begin{bmatrix}
    10 & 3 \\
  \end{bmatrix},
\end{equation}
and the Lotka--Volterra equations
\begin{equation}
  \begin{cases}
    \dot u &= (a - bv) u \\
    \dot v &= (du - c) v
    \label{equation:lotkavolterra-ivp}
  \end{cases}
  \quad\quad\quad\quad
  t_0 = 0
  \quad\quad
  \begin{cases}
    u(t_0) &= \frac{1}{2} \\
    v(t_0) &= \frac{1}{2}
  \end{cases}
  \quad\quad
  \begin{bmatrix}
    a & b \\
    c & d 
  \end{bmatrix}
  =
  \begin{bmatrix}
    10 & 10 \\
    10 & 10 
  \end{bmatrix}.
\end{equation}

\begin{figure}[H]
  \begin{center}
    \includegraphics[width=\linewidth]{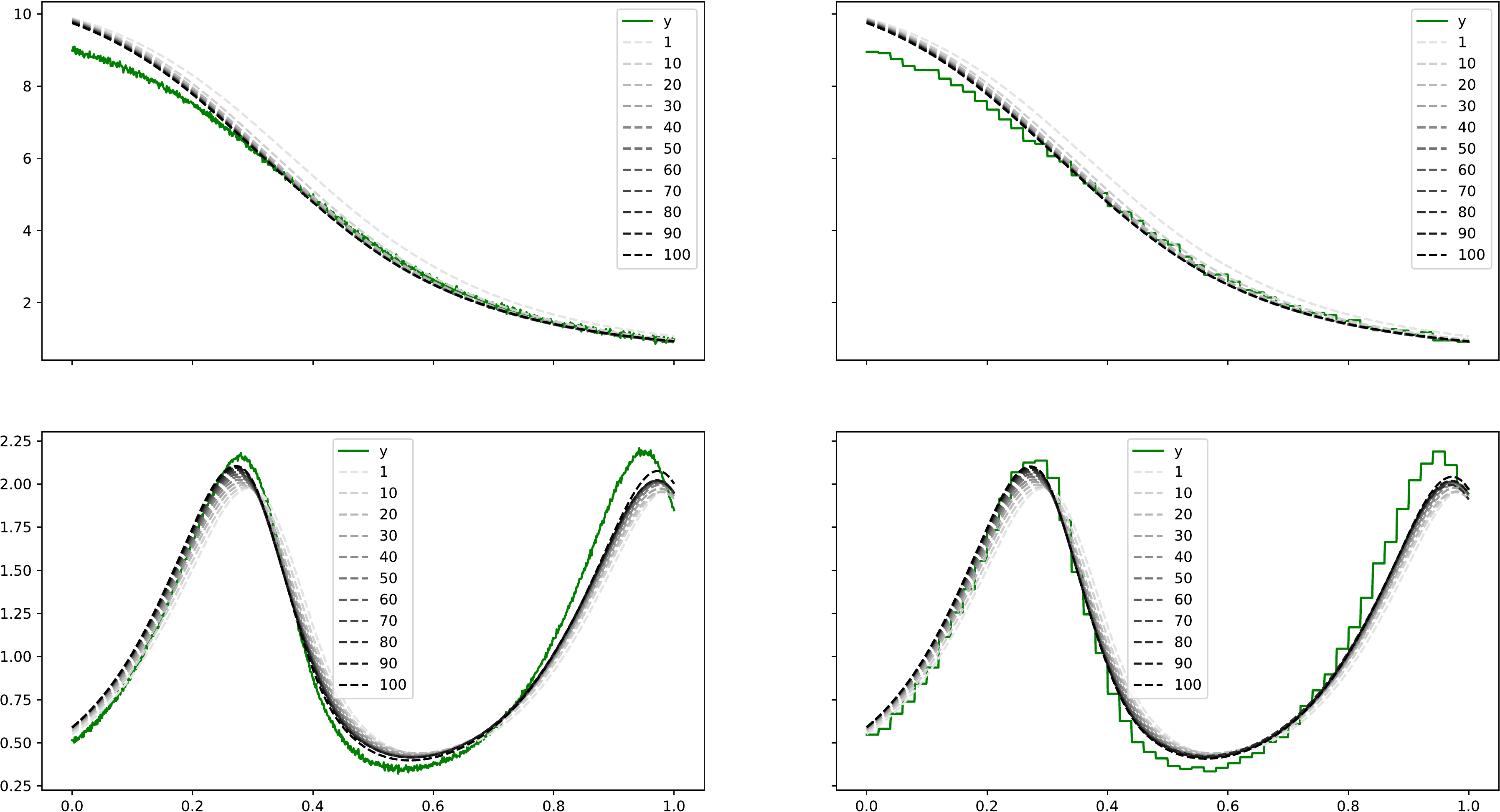}
    \\
    \includegraphics[width=\linewidth]{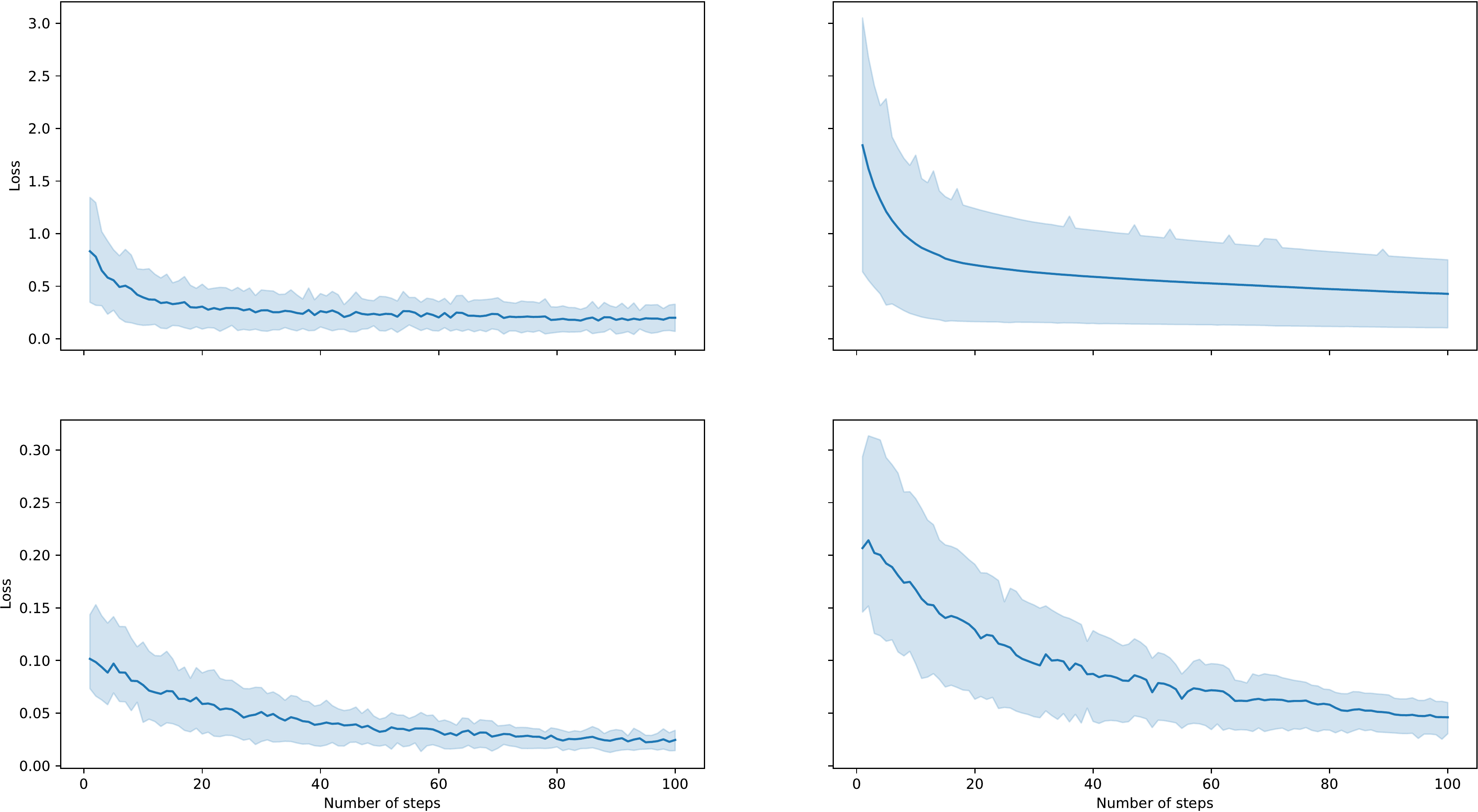}
  \end{center}
  \caption{
    The two quadruples depict the results of $100$ gradient descent steps starting from a slightly perturbed initial value problem parameter triple $(t_0, x_0, \theta)$.
    In each quadruple, the first row belongs to the case of the SI model
    \Cref{equation:sir-ivp}, while the second to that of the Lotka--Volterra equations \Cref{equation:lotkavolterra-ivp}. The first column shows the case of continuous input,
    the second that of discrete input.
    The upper quadruple shows
    the input data $y$, and how the current best estimate of the underlying
    trajectory component varies during the iteration,
    The lower quadruple shows 
    the loss values encountered during the same time.
    The latter are based on $4$ repetitions of each experiment.
  }
\label{figure:experiments-losses-progress}
\end{figure}

We have ran the experiment for each set of input data, for each initial value
problem. We have repeated each experiment $4$ times, so as to get a better idea of the loss values encountered during the iteration. The results of the $2\times 2 \times 4$ experiments
are summarized in \Cref{figure:experiments-losses-progress}.

The experiments have been implemented in JAX \cite{JAX}.
The implementation tries to mimic the mathematics presented in this paper.
In particular, it has not been optimized for computational efficiency.
In practice, calculating the gradients requires the numerical solution of
an initial value problem, and further numerical integration. 
This implies that the amount of work required for each gradient descent step
depends on the numerical tolerances one specifies, with looser tolerances
implying faster iteration. On the other hand, looser tolerances imply less precise
gradients. It is unclear how these tolerances should be chosen, perhaps even
varied during the iteration, to render the computational process more efficient
in terms of the decrement of the loss value per unit work.

In the continuous case, increasing the amount of noise, the integrals become harder to evaluate, which
results in increased computation time and decreased accuracy.
In the discrete case, taking samples from each subinterval according to a
truncated normal distribution implies that as the temporal uncertainty goes to
zero, the value of the weight function at the midpoints goes to infinity, which corresponds
to the discrete part of \Cref{equation:adjoint-ivp}.

The evaluation of the loss function, that is, that of the final integral, is not necessary for the calculation of the gradients, and time may be saved by only evaluating it when necessary.

In the examples of this section, the parameter triple the gradient descent
starts from is not far from the one which yields the input data.
When the initial parameter triple is further, then the true and the predicted
trajectories can be different enough qualitatively for the iterative process
to get stuck. In these cases, one may mimic the idea of the stochastic gradient
descent by replacing $\sigma$ with a random measure for each gradient descent step.
We have had success using random normal distributions that were modified so
that the expected measure was approximately uniform on the unit interval.
This uniformity appears important in making sure that on average, the
stochastic choice of measure does not interfere with how the errors at each time
instant are weighted.

\section*{Funding}
I.F. was supported by the János Bolyai Research Scholarship of the Hungarian Academy of Sciences.

This research has been implemented with the support provided by the Ministry of Innovation and Technology of Hungary from the National Research, Development and Innovation Fund, financed under the ELTE TKP 2021-NKTA-62 funding scheme.

P.L.S. acknowledges support from the Hungarian Scientific Research Fund, OTKA (grant no. 135241) and from the Ministry of Innovation and Technology NRDI Office within the framework of the Artificial Intelligence National Laboratory Programme.

\end{document}